\documentclass{article}

\usepackage{arxiv}
\usepackage[utf8]{inputenc}
\usepackage[T1]{fontenc}
\usepackage{hyperref}
\usepackage{url}
\usepackage{booktabs}
\usepackage{amsfonts}
\usepackage{nicefrac}
\usepackage{microtype}
\usepackage{graphicx}
\usepackage{natbib}
\usepackage{doi}
\usepackage{amsmath}
\usepackage{amssymb}
\usepackage{mathtools}
\usepackage{subcaption}
\usepackage{float}

\title{SciML in the Wild: A Diagnostic Study of When Structural Priors Help and When They Hurt}

\author{
Vrishank Sai Anand \\
GEMS Modern Academy \\
Dubai, United Arab Emirates \\
\texttt{vrishanksai.anand@gmail.com} \\
\And
 Prathamesh Dinesh Joshi \\
  Vizuara AI Labs,\\
  Pune, India \\
  \texttt{prathamesh@vizuara.com} \\
   \And
 Raj Abhijit Dandekar \\
  Vizuara AI Labs\\
  Pune, India\\
  \texttt{raj@vizuara.com} \\
   \And
 Rajat Dandekar\\
   Vizuara AI Labs\\
  Pune, India\\
  \texttt{rajatdandekar@vizuara.com} \\
   \And
 Sreedath Panat\\
  Vizuara AI Labs\\
  Pune, India\\
  \texttt{sreedath@vizuara.com} \\
}
\begin{document}
\maketitle

\begin{abstract}
Scientific Machine Learning (SciML) methods such as Neural Ordinary Differential Equations (NODEs), Physics-Informed Neural Networks (PINNs), and Universal Differential Equations (UDEs) are most effective when structural priors reflect reliable governing dynamics. We ask what happens when this assumption is violated. Using macroeconomic forecasting as a stress-test domain, we evaluate five model families, ARIMA, LSTM, NODE, PINN, and UDE, across 23 countries using sparse annual data, multiple temporal splits, and five random seeds. Our results show that none of the evaluated models achieve consistently strong forecasting performance, highlighting the difficulty of low-frequency macroeconomic prediction. However, a clear relative hierarchy emerges: less-constrained models, particularly ARIMA and NODE, consistently outperform more-constrained heuristic-prior models such as PINN and UDE. Rather than treating this as a rejection of SciML, we interpret it as a diagnostic result: structural priors can act as misregularizers when they do not match the data-generating process. We identify failure modes including prior misalignment, regime shifts, structural breaks, and optimization instability, and argue that SciML practitioners should test whether structure helps before assuming that more structure is beneficial.
\end{abstract}


\section{Introduction}

Macroeconomic crises rarely emerge as isolated shocks. Instead, they develop through the interaction of debt accumulation, growth dynamics, inflation, and policy responses over extended periods. Empirical evidence from international finance shows that such crises are often preceded by persistent macroeconomic imbalances and systemic vulnerabilities \citep{kaminsky1999twin, berg1999predicting, caballero2021global}. Despite the presence of these recurring patterns, forecasting macroeconomic dynamics remains a difficult problem, particularly in low-frequency and data-constrained settings.

Traditional approaches to macroeconomic forecasting, including econometric models such as ARIMA, VAR, and Early Warning Systems, rely on statistical relationships between macroeconomic indicators \citep{box2016time, lutkepohl2005new, hamilton1994time, berg1999predicting}. While these methods provide interpretable baselines, they often struggle to capture nonlinear interactions, structural breaks, and regime shifts that are common in real-world economies \citep{hamilton1989new, makridakis2018statistical}. More recent machine learning approaches, including sequence models such as LSTMs, attempt to address these limitations by learning patterns directly from data \citep{hochreiter1997long, goodfellow2016deep}, but they remain sensitive to data scarcity, non-stationarity, and distributional instability \citep{makridakis2018statistical, lim2021time}. Recent advances in deep forecasting, including transformer-based architectures, have shown strong performance in large-scale benchmarks, but their applicability to macroeconomic data remains limited due to data sparsity and regime instability \citep{zeng2023transformers, liu2022non}.

Scientific Machine Learning (SciML) offers an alternative paradigm by incorporating structural priors into learning. Methods such as Neural Ordinary Differential Equations (NODEs) \citep{chen2018neural}, Physics-Informed Neural Networks (PINNs) \citep{raissi2019physics}, and Universal Differential Equations (UDEs) \citep{rackauckas2020universal} aim to combine data-driven learning with known system dynamics. These approaches have demonstrated strong performance in domains such as physics, biology, and engineering, where the governing equations are well-defined and stable \citep{karniadakis2021physics, brunton2022data}. In these settings, structural constraints act as inductive biases that improve generalization and reduce data requirements.

However, the applicability of these methods to macroeconomics is unclear. Unlike physical systems, macroeconomic environments are shaped by human behavior, institutional decisions, and external shocks \citep{farmer2009economy}. As a result, the underlying dynamics are often non-stationary, regime-dependent, and only partially observable \citep{hamilton1989new, farmer2009economy}. This raises a fundamental question: when structural priors are uncertain or misaligned with the data-generating process, do they improve model performance or can they degrade it?

In this work, we address this question through a large-scale empirical study in 23 countries. We treat macroeconomic forecasting as a stress-test domain for SciML. The data is sparse, low-frequency, noisy, and subject to structural breaks. We compare five model families, ARIMA, LSTM, Neural ODE (NODE), PINN, and UDE, under a unified evaluation framework using multiple temporal splits and repeated random initialization. Importantly, the physics-informed components used in this study are heuristic structural priors rather than theory-derived economic models. This allows us to isolate the effect of inductive constraints in a controlled setting.

This study investigates whether structural priors improve forecasting performance in sparse, non-stationary macroeconomic settings, or whether they can become harmful when misaligned with the underlying dynamics \citep{hamilton1989new, makridakis2018statistical, farmer2009economy}. To answer this, we conduct a large-scale empirical evaluation across 23 countries, comparing classical statistical baselines, sequence models, continuous-time neural dynamics, and structurally constrained SciML approaches under a unified experimental framework \citep{box2016time, hochreiter1997long, chen2018neural, raissi2019physics, rackauckas2020universal}.

Rather than assuming that additional structural bias is necessarily beneficial, we examine how different levels of imposed structure interact with noisy, low-frequency, and regime-dependent macroeconomic data. Our goal is not to evaluate theory-grounded economic structural models, but to diagnose the behavior of heuristic structural priors when applied outside domains with well-specified governing equations \citep{karniadakis2021physics, brunton2022data}.

The contributions of this work are as follows:

\begin{itemize}
    \item We provide a 23-country empirical evaluation of SciML methods in macroeconomic forecasting, a domain characterized by sparse, non-stationary, and behavior-driven data.
    
    \item We show that heuristic structural priors in PINN and UDE models do not improve, and often degrade, predictive performance relative to less-constrained approaches such as ARIMA and NODE.
    
    \item We identify and characterize four failure modes of physics-informed learning under non-stationarity, providing a diagnostic framework for when structural priors may harm rather than help.
\end{itemize}

These results suggest that the effectiveness of SciML depends critically on the validity of its underlying assumptions. In domains where governing equations are uncertain or evolving, flexibility may be more valuable than imposed structure. This highlights the importance of testing inductive priors empirically rather than assuming their benefit a priori.

\section{Related Work}

\paragraph{Macroeconomic forecasting.}
Macroeconomic forecasting has traditionally relied on statistical and econometric models such as ARIMA, VAR, and factor-based approaches \citep{box2016time, lutkepohl2005new, hamilton1994time}. Early Warning Systems have also been widely studied for predicting financial crises using macroeconomic indicators \citep{kaminsky1999twin, berg1999predicting}. More recent work has extended these approaches using machine learning methods to improve macro-financial modeling and forecasting \citep{gu2020empirical}. While these methods provide interpretable baselines, their predictive performance remains limited in out-of-sample settings, particularly in the presence of structural breaks and regime shifts \citep{hamilton1989new, caballero2021global}. These limitations motivate the exploration of more flexible modeling approaches.

\paragraph{Machine learning for time series forecasting.}
Recent advances in machine learning have introduced data-driven approaches for time series forecasting, including recurrent neural networks and attention-based architectures \citep{hochreiter1997long, goodfellow2016deep}. More recently, transformer-based and hybrid architectures have shown strong performance in large-scale forecasting benchmarks \citep{lim2021time, zeng2023transformers, liu2022non}. However, their effectiveness in macroeconomic settings remains constrained by limited data availability, low sampling frequency, and non-stationarity \citep{makridakis2018statistical}. As a result, purely data-driven models often exhibit overfitting and instability when applied to macroeconomic time series.

\paragraph{Scientific machine learning and physics-informed modeling.}
Scientific Machine Learning (SciML) introduces an alternative paradigm by incorporating structural priors into learning. Methods such as Neural Ordinary Differential Equations (NODEs) \citep{chen2018neural}, Physics-Informed Neural Networks (PINNs) \citep{raissi2019physics}, and Universal Differential Equations (UDEs) \citep{rackauckas2020universal} have demonstrated strong performance in domains where governing equations are known or well-approximated \citep{karniadakis2021physics, brunton2022data}. These approaches use structural constraints as inductive biases to improve generalization and data efficiency. However, their application outside physical sciences remains relatively underexplored. In particular, few studies systematically evaluate how such structural priors behave in domains characterized by non-stationarity, regime shifts, and partially observed dynamics.

\paragraph{Positioning of this work.}
This paper contributes to this gap by treating macroeconomic forecasting as a stress-test domain for SciML. Unlike prior work that assumes the validity of structural priors, we explicitly evaluate their behavior when such assumptions are uncertain or misaligned with the data-generating process. Rather than proposing a new model, our contribution is diagnostic: we provide empirical evidence across 23 countries showing when and why structural priors can degrade performance, and we identify failure modes that are critical for applying SciML methods in non-physical domains.

\section{Methodology}

\subsection{Data and State Representation}

We construct a multi-country macroeconomic dataset using publicly available data from the World Bank and IMF databases. The dataset consists of annual observations for 23 countries spanning multiple regions and economic regimes. Depending on availability, individual country series cover either the period 1990–2024 ($N=35$) or 1960–2024 ($N=65$).

For each country, the macroeconomic system at time $t$ is represented by a five-dimensional state vector:

\begin{equation}
\mathbf{u}(t) =
\begin{bmatrix}
\text{GDP}_{\text{USD}}(t) \\
\dot{Y}(t) \\
\pi(t) \\
D(t) \\
E(t)
\end{bmatrix}
\in \mathbb{R}^5,
\end{equation}

where $\text{GDP}_{\text{USD}}$ denotes gross domestic product in current U.S. dollars, $\dot{Y}$ is real GDP growth rate (\%), $\pi$ is inflation (\%), $D$ is government debt as a percentage of GDP, and $E$ is external debt in U.S. dollars. These variables jointly capture economic scale, growth dynamics, price stability, fiscal stress, and external vulnerability.

To ensure comparability across variables with different magnitudes, all features are normalized to the interval $[0,1]$ using MinMax scaling. To prevent data leakage, the scaler is fitted exclusively on the training portion of each country-level time series and then applied to validation and test data.

\subsection{Temporal Split Protocol}

To preserve the sequential structure of macroeconomic data, all datasets are split chronologically. For a country with $N$ observations, we define:

\begin{equation}
N_{\text{train}} = \lfloor 0.70N \rfloor, \quad
N_{\text{val}} = \lfloor 0.15N \rfloor, \quad
N_{\text{test}} = N - N_{\text{train}} - N_{\text{val}}.
\end{equation}

In addition to this default split, robustness checks are performed using alternative partitions of 65/15/20 and 80/10/10.

For sequence-based models, including LSTM and PINN, inputs are constructed using sliding windows of length $L=5$. To maintain strict temporal separation, test sequences are initialized using the final $L$ observations from the validation set, ensuring that no test data is included within any training input window.

\paragraph{On structural priors.}
The structural constraints used in this study are heuristic and not derived from formal macroeconomic theory such as DSGE or RBC models. Instead, they are simplified assumptions designed to test the effect of embedding structure into learning. This distinction is important: our goal is not to evaluate physics-informed modeling in its full theoretical form, but to examine how generic structural priors behave when applied to macroeconomic data. As a result, our conclusions should be interpreted as a diagnostic of prior misalignment rather than a general critique of theory-driven modeling.

\subsection{Modeling Approaches}

We evaluate five modeling paradigms that span classical statistical forecasting, data-driven sequence learning, continuous-time dynamics, and physics-informed learning.

\paragraph{ARIMA.}
As a statistical baseline, we fit a univariate ARIMA(1,1,1) model independently for each macroeconomic indicator. The model is trained on the combined training and validation data and evaluated using one-step-ahead rolling forecasts over the test period. Due to its deterministic nature, ARIMA does not require averaging across random seeds.

\paragraph{LSTM.}
To capture temporal dependencies directly from data, we use a sequence-based LSTM model. Given a window of past observations, the model predicts the next macroeconomic state:

\begin{equation}
\hat{\mathbf{u}}(t+1) = f_{\text{LSTM}}\big(\mathbf{u}(t-L+1:t)\big).
\end{equation}

The architecture consists of a two-layer LSTM with hidden size 64 and dropout 0.15, followed by a linear output layer. The model is trained using the Adam optimizer with learning rate $\eta = 8 \times 10^{-4}$ and weight decay $10^{-3}$, with early stopping based on validation loss.

\paragraph{Neural ODE (NODE).}
To model macroeconomic dynamics in continuous time, we use a Neural Ordinary Differential Equation defined as:

\begin{equation}
\frac{d\mathbf{u}}{dt} = f_\theta(\mathbf{u}),
\end{equation}

where $f_\theta : \mathbb{R}^5 \rightarrow \mathbb{R}^5$ is a neural network parameterizing the system dynamics. The vector field is implemented as a two-layer multilayer perceptron with 32 hidden units and Tanh activations.

The system is numerically integrated using a fixed-step Runge–Kutta (RK4) solver. To improve stability on sparse annual data, we employ a multiple-shooting training strategy:

\begin{equation}
\mathcal{L}_{\text{NODE}} =
\frac{1}{K} \sum_{k=1}^{K}
\left\|
\Phi_{t_k \rightarrow t_k+S}(\mathbf{u}_{t_k})
- \mathbf{u}_{t_k:t_k+S}
\right\|^2
+ \lambda_c \mathcal{L}_{\text{cont}},
\end{equation}

where $\Phi$ denotes the numerical flow map, and $\mathcal{L}_{\text{cont}}$ penalizes discontinuities between segments.

\paragraph{PINN (heuristic structural priors).}
The PINN model augments data-driven learning with soft structural constraints. The total loss is defined as:

\begin{equation}
\mathcal{L}_{\text{PINN}} = \mathcal{L}_{\text{data}} + \lambda_\phi \mathcal{L}_{\text{prior}},
\end{equation}

where $\mathcal{L}_{\text{prior}}$ encodes heuristic relationships between macroeconomic variables:

\begin{equation}
\mathcal{L}_{\text{prior}} =
\frac{1}{4} \sum_{i=1}^{4} \|\mathcal{R}_i\|^2.
\end{equation}

The residuals $\mathcal{R}_i$ represent simplified assumptions including debt-growth interaction, growth mean-reversion, inflation–debt coupling, and GDP smoothness. These residuals are computed using finite-difference approximations on predicted trajectories.

Unlike standard PINN formulations that rely on automatic differentiation and known governing equations \citep{raissi2019physics}, our implementation uses finite differences due to the discrete and low-frequency nature of macroeconomic data. This design choice reflects the absence of a continuous and well-specified underlying system, and allows us to evaluate how approximate structural priors interact with data-driven learning.

\paragraph{UDE (hybrid structural model).}
The UDE combines a predefined structural component with a learned residual:

\begin{equation}
\frac{d\mathbf{u}}{dt} = f_{\text{known}}(\mathbf{u}) + f_\theta(\mathbf{u}).
\end{equation}

The known component encodes simplified macroeconomic relationships, while the neural component captures deviations from this structure. To avoid overwhelming the learned dynamics, the residual network is initialized with small weights so that training begins close to the predefined system.

As with the PINN formulation, the structural component is intuitive rather than theory-derived. This allows us to evaluate whether partially specified dynamics improve forecasting performance in the presence of limited and noisy data.

\subsection{Evaluation Protocol}

All models are evaluated on the normalized scale to ensure a fair comparison across countries and features. The primary metric is the coefficient of determination ($R^2$), computed as
\begin{equation}
R^2
=
1-
\frac{\sum_t (u_t-\hat{u}_t)^2}
{\sum_t (u_t-\bar{u})^2}.
\end{equation}
We report the mean $R^2$ across all five features for each country. For stochastic models (LSTM, NODE, UDE, and PINN), results are averaged over five random seeds $\{42, 7, 123, 2024, 999\}$ and reported as mean $\pm$ standard deviation.

For continuous-time models, including NODE and UDE, evaluation is performed in a one-step-ahead setting: given the observed state $\mathbf{u}(t)$, the learned dynamics are integrated forward by one year to predict $\mathbf{u}(t+1)$. This design keeps the forecasting task comparable across models while still allowing continuous-time methods to operate in their natural dynamical setting.

\subsection{Leakage Prevention}

Because macroeconomic time series are short and highly structured, preventing leakage is essential. Three safeguards are enforced throughout the pipeline. First, scaler leakage is prevented by fitting the MinMax scaler only on the training segment. Second, sequence boundary leakage is avoided by constructing validation and test windows so that no future target appears in any input window. Third, hyperparameter leakage is controlled by fixing model settings before test evaluation and using validation performance only for early stopping and model selection. These measures ensure that all reported results reflect genuine out-of-sample behavior rather than information inadvertently transferred from future data.

We note that macroeconomic forecasting under these conditions is inherently difficult, and many models evaluated in this study achieve negative $R^2$ values, indicating performance below naive baselines. Our focus is therefore on relative model behavior rather than absolute predictive success.

\section{Results}

In this section, we evaluate the performance of all models across 23 countries using multiple temporal splits. We focus on both predictive accuracy and model stability, with particular attention to how different modeling paradigms behave under low-data macroeconomic conditions. Results are analyzed through quantitative metrics ($R^2$, RMSE, MAPE) as well as qualitative trajectory reconstruction.

\begin{table}[t]
\centering
\small
\resizebox{\columnwidth}{!}{
\begin{tabular}{lccccc}
\toprule
\textbf{Country} & \textbf{LSTM} & \textbf{NeuralODE} & \textbf{UDE} & \textbf{PINN} & \textbf{ARIMA} \\
\midrule
Argentina & -5.911 $\pm$ 2.351 & -0.048 $\pm$ 0.056 & -2.549 $\pm$ 0.989 & -50.094 $\pm$ 36.792 & \textbf{0.174} \\
Bangladesh & -20.703 $\pm$ 3.363 & -4.242 $\pm$ 0.274 & -28.874 $\pm$ 1.328 & -72.081 $\pm$ 18.909 & \textbf{0.157} \\
Brazil & -161.075 $\pm$ 154.633 & -942.878 $\pm$ 522.839 & -619.443 $\pm$ 85.492 & -33.865 $\pm$ 16.607 & \textbf{-0.521} \\
Chile & -7.871 $\pm$ 0.807 & -0.662 $\pm$ 0.043 & -4.102 $\pm$ 0.113 & -5.603 $\pm$ 0.578 & \textbf{-0.576} \\
Colombia & -6.932 $\pm$ 1.382 & -0.432 $\pm$ 0.005 & -2.039 $\pm$ 0.076 & -5.700 $\pm$ 0.698 & \textbf{-0.155} \\
Egypt & -9.366 $\pm$ 1.758 & -1.333 $\pm$ 0.155 & -6.280 $\pm$ 0.575 & -52.089 $\pm$ 8.755 & \textbf{-0.654} \\
Greece & -0.882 $\pm$ 0.254 & -0.311 $\pm$ 0.006 & -1.598 $\pm$ 0.058 & -1.452 $\pm$ 0.593 & \textbf{-0.234} \\
India & -9.053 $\pm$ 0.468 & \textbf{0.052 $\pm$ 0.008} & -12.344 $\pm$ 0.970 & -171.906 $\pm$ 40.411 & -0.078 \\
Indonesia & -37.697 $\pm$ 5.708 & -24.198 $\pm$ 2.852 & -9.708 $\pm$ 1.326 & -21.476 $\pm$ 8.043 & \textbf{-2.194} \\
Italy & -2.790 $\pm$ 1.227 & -0.571 $\pm$ 0.001 & -6.792 $\pm$ 0.245 & -1.981 $\pm$ 0.524 & \textbf{-0.505} \\
Japan & -15.787 $\pm$ 1.164 & -0.052 $\pm$ 0.027 & -0.484 $\pm$ 0.008 & -13.394 $\pm$ 0.553 & \textbf{-0.034} \\
Mexico & -31.446 $\pm$ 24.319 & -0.332 $\pm$ 0.006 & -4.316 $\pm$ 0.078 & -17.022 $\pm$ 12.775 & \textbf{-0.140} \\
Nigeria & -15.849 $\pm$ 2.723 & 0.033 $\pm$ 0.005 & -0.698 $\pm$ 0.355 & -6.645 $\pm$ 0.651 & \textbf{0.046} \\
Pakistan & -10.924 $\pm$ 1.517 & -1.865 $\pm$ 0.368 & -6.288 $\pm$ 0.514 & -11.789 $\pm$ 2.442 & \textbf{-0.524} \\
Peru & -534.517 $\pm$ 580.080 & -66.705 $\pm$ 15.173 & -11638.053 $\pm$ 3219.406 & -182.678 $\pm$ 180.409 & \textbf{-0.188} \\
Philippines & -2.118 $\pm$ 0.177 & -2.452 $\pm$ 0.146 & -7.423 $\pm$ 0.738 & -1.051 $\pm$ 0.511 & \textbf{0.174} \\
Romania & -8.075 $\pm$ 5.288 & -6.496 $\pm$ 0.286 & -6.413 $\pm$ 0.269 & -5.690 $\pm$ 1.258 & \textbf{-0.047} \\
South Africa & -27.842 $\pm$ 4.020 & -3.571 $\pm$ 0.117 & -5.784 $\pm$ 0.107 & -24.368 $\pm$ 3.876 & \textbf{-2.258} \\
Sri Lanka & -23.922 $\pm$ 0.808 & -0.110 $\pm$ 0.015 & -5.570 $\pm$ 0.047 & -21.911 $\pm$ 2.750 & \textbf{0.024} \\
Thailand & -2.873 $\pm$ 0.744 & -2.288 $\pm$ 0.192 & -11.560 $\pm$ 1.116 & -2.046 $\pm$ 0.457 & \textbf{-0.717} \\
Turkey & -1.048 $\pm$ 0.177 & -0.105 $\pm$ 0.002 & -0.103 $\pm$ 0.020 & -0.548 $\pm$ 0.114 & \textbf{0.231} \\
Ukraine & -53.552 $\pm$ 76.652 & -357.664 $\pm$ 33.126 & -176.287 $\pm$ 32.584 & -10.249 $\pm$ 15.770 & \textbf{-0.556} \\
Vietnam & -65.991 $\pm$ 4.771 & -0.844 $\pm$ 0.132 & -8.625 $\pm$ 1.611 & -36.396 $\pm$ 16.781 & \textbf{-0.048} \\
\bottomrule
\end{tabular}
}
\caption{Comparison of model performance across 23 countries under the 70/15/15 temporal split. Values report mean $R^2$ $\pm$ standard deviation over five random seeds. The best-performing model for each country is highlighted in bold.}
\label{tab:main_results}
\end{table}

\subsection{Main Comparison}

Table~\ref{tab:main_results} presents the primary results across all 23 countries under the 70/15/15 temporal split, reporting mean $R^2$ $\pm$ standard deviation over five random seeds.

A first and important observation is that macroeconomic forecasting in this setting is intrinsically difficult. Across all models and countries, the majority of $R^2$ values are negative, indicating that most approaches perform worse than a naive baseline that predicts the mean. This highlights the challenge posed by sparse annual data, non-stationarity, and structural breaks.

\begin{figure}[t]
    \centering
    \includegraphics[width=0.8\columnwidth]{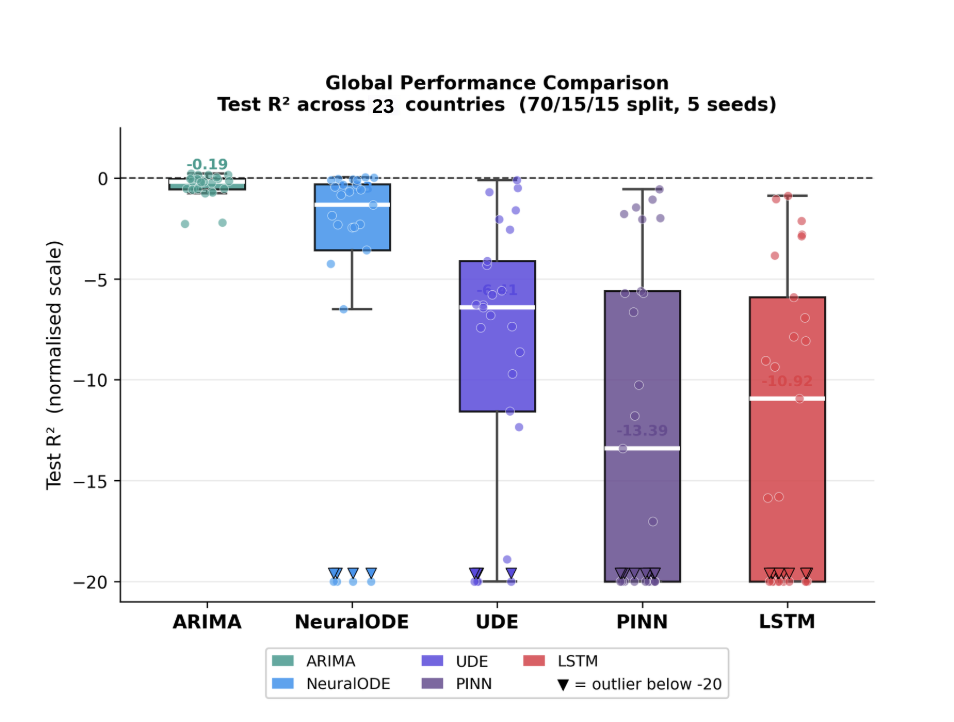}
    \caption{Distribution of test $R^2$ across 23 countries. ARIMA and NODE achieve higher median performance, while PINN and UDE show lower medians and higher variability. Values below $-20$ are clipped for visualization.}
    \label{fig:global_performance}
\end{figure}

As shown in Figure~\ref{fig:global_performance}, this difficulty is not confined to isolated failure cases, but appears systematically across model families. The distribution of test $R^2$ values reveals that ARIMA and Neural ODE (NODE) achieve comparatively higher median performance with narrower spread, while structurally constrained models such as PINN and UDE exhibit both lower central performance and substantially greater variance. The presence of multiple clipped outliers below $-20$ further suggests catastrophic failures in certain country-model combinations, particularly for heavily constrained architectures.

Despite this, a consistent relative hierarchy emerges across model families. Less-constrained models, particularly ARIMA and Neural ODE (NODE), outperform more-constrained approaches such as PINN and UDE in the majority of countries. This pattern is robust across both stable and volatile macroeconomic regimes.

The dominance of ARIMA should be interpreted carefully. Under the one-step-ahead evaluation protocol, ARIMA benefits from conditioning directly on the most recent observed value at each timestep. This creates a structural advantage in short test windows. However, the more informative comparison is within the neural model family. Here, NODE consistently outperforms both PINN and UDE, despite not enforcing explicit structural constraints.

This result is central to our study. While PINN and UDE introduce structural priors intended to regularize learning, these constraints do not improve performance in practice. Instead, they often degrade predictive accuracy, particularly in countries experiencing volatility, crisis events, or regime shifts. In such settings, the imposed priors are frequently misaligned with the underlying data-generating process.

Importantly, the underperformance of structured models is not limited to accuracy alone. As shown in Table~\ref{tab:main_results}, PINN exhibits both large negative $R^2$ values and high variance across random seeds, indicating unstable optimization. UDE shows a similar but less extreme pattern. This suggests that the introduction of structural constraints can increase sensitivity to initialization and amplify training instability.

Taken together, these results support a key conclusion: in macroeconomic systems characterized by sparse, noisy, and evolving dynamics, structural priors do not necessarily improve learning. When the assumed structure is incorrect or incomplete, it can act as a source of bias rather than a useful inductive guide. In contrast, models that preserve flexibility, such as NODE, are better able to adapt to observed data, even under limited training conditions.

\subsection{Uncertainty and Stability Analysis}

\begin{figure*}[t]
    \centering
    \includegraphics[width=\textwidth]{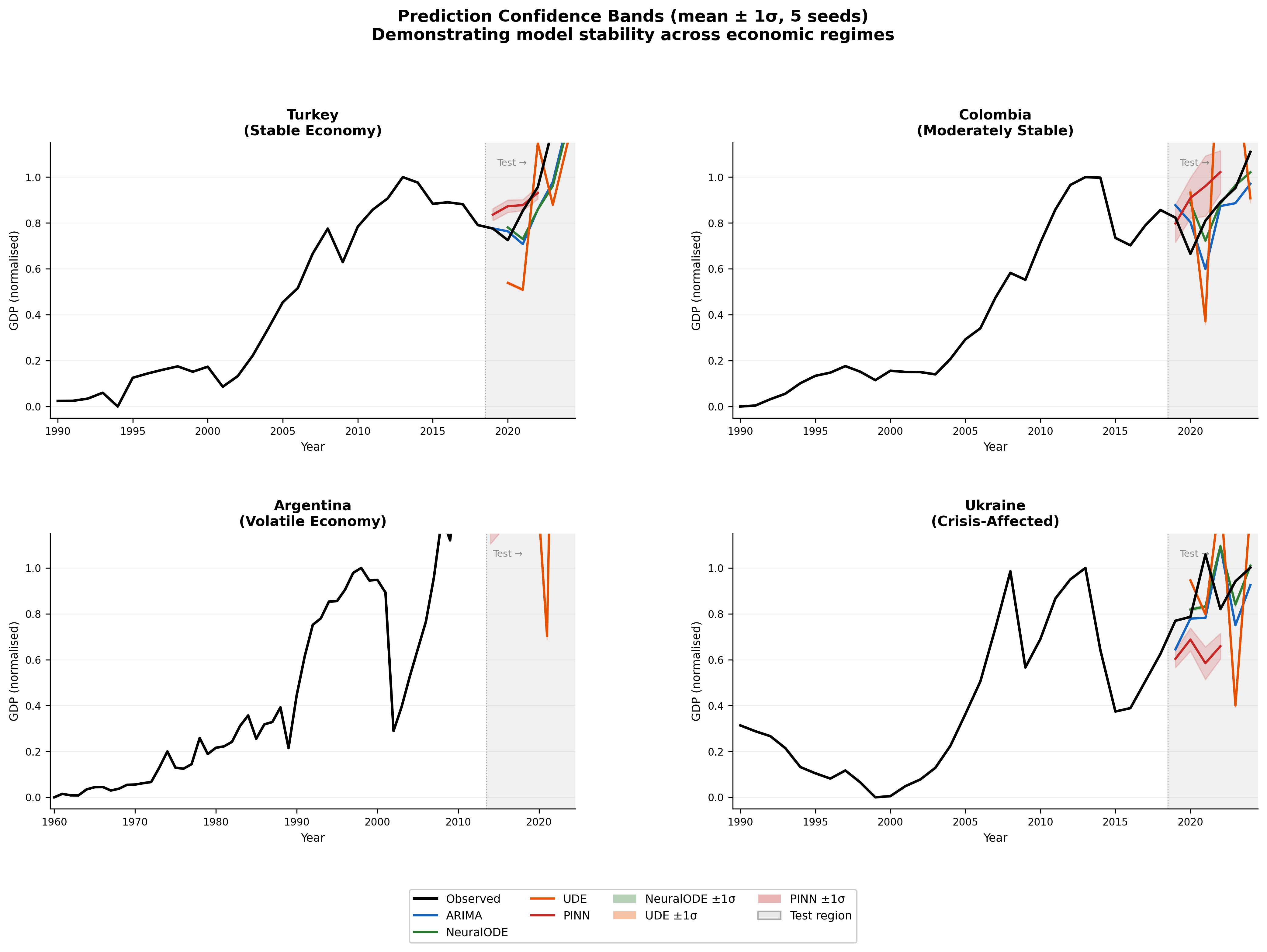}
    \caption{Prediction confidence bands (mean $\pm$ 1 standard deviation across five seeds) for representative economies. NODE exhibits consistently narrow bands, indicating stable dynamics, while PINN shows wide bands in volatile regimes, reflecting instability. UDE is intermediate and ARIMA is deterministic.}
    \label{fig:confidence_bands}
\end{figure*}

Figure~\ref{fig:confidence_bands} evaluates model stability by showing prediction confidence bands across five random seeds for representative countries. This analysis complements the aggregate metrics by assessing reproducibility and sensitivity to initialization.

A clear pattern emerges across models. ARIMA produces deterministic predictions with zero variance, while NODE exhibits consistently narrow confidence bands, indicating stable and reproducible dynamics. In contrast, models with structural priors show significantly higher uncertainty. PINN produces the widest bands, particularly in volatile and crisis-affected economies, while UDE exhibits intermediate variability.

This behavior reflects the interaction between structural constraints and data. When the imposed priors are approximately aligned with the observed dynamics, their effect is limited. However, in the presence of structural breaks or regime changes, these constraints introduce gradients that conflict with the data, leading to unstable training and divergence across random seeds.

The key insight is that structural priors do not automatically improve robustness. In this setting, they often increase variance rather than reduce it. This provides further evidence that the effectiveness of inductive bias depends critically on its alignment with the underlying system.

\subsection{Robustness Across Temporal Splits}

\begin{figure*}[t]
    \centering
    \includegraphics[width=\textwidth]{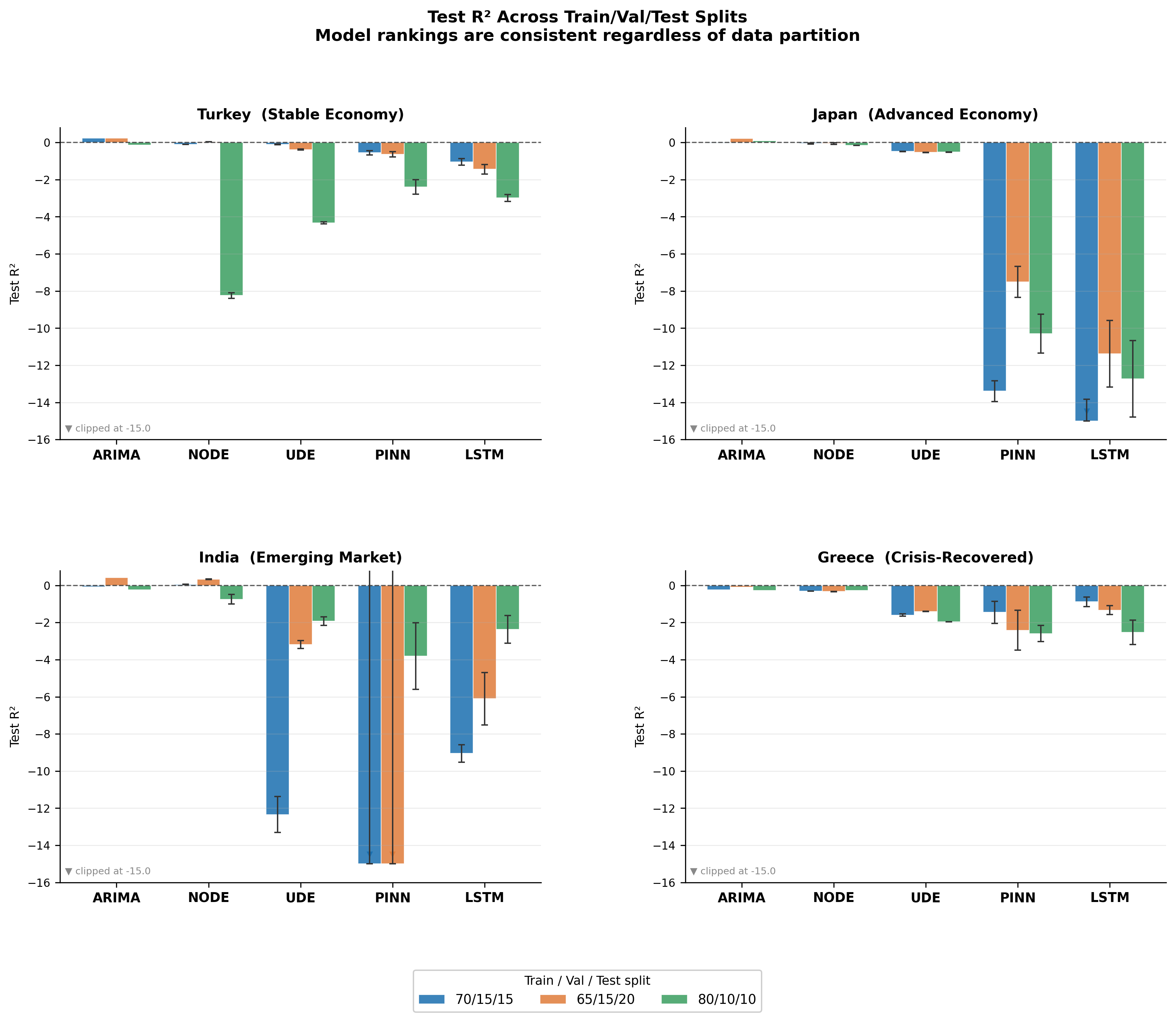}
    \caption{Test $R^2$ across different temporal splits (70/15/15, 65/15/20, 80/10/10). Model rankings remain consistent, with ARIMA and NODE outperforming PINN and UDE, indicating robustness to split choice.}
    \label{fig:split_sensitivity}
\end{figure*}

Figure~\ref{fig:split_sensitivity} examines whether the observed model behavior depends on the choice of temporal split. Across all evaluated configurations, the relative ranking of model families remains consistent.

Less-constrained models, particularly ARIMA and NODE, continue to outperform more-constrained approaches such as PINN and UDE. While absolute $R^2$ values vary as the size of the training set changes, the overall hierarchy is preserved.

This consistency is important because it rules out the possibility that the results are driven by a particular train-test boundary. Instead, the observed patterns reflect systematic differences in how model assumptions interact with macroeconomic data characteristics.

In particular, the persistent underperformance of PINN and UDE across splits suggests that their limitations are structural rather than incidental. When the imposed priors do not match the underlying dynamics, increasing or decreasing the amount of training data does not resolve the mismatch.

\subsection{Ablation of Structural Priors}

To determine whether the poor performance of physics-informed models arises from the neural architecture itself or from the imposed structural constraints, we perform an ablation study on the PINN formulation by removing the physics-informed loss term while keeping the underlying predictive architecture unchanged.

\begin{figure*}[t]
    \centering
    \includegraphics[width=\textwidth]{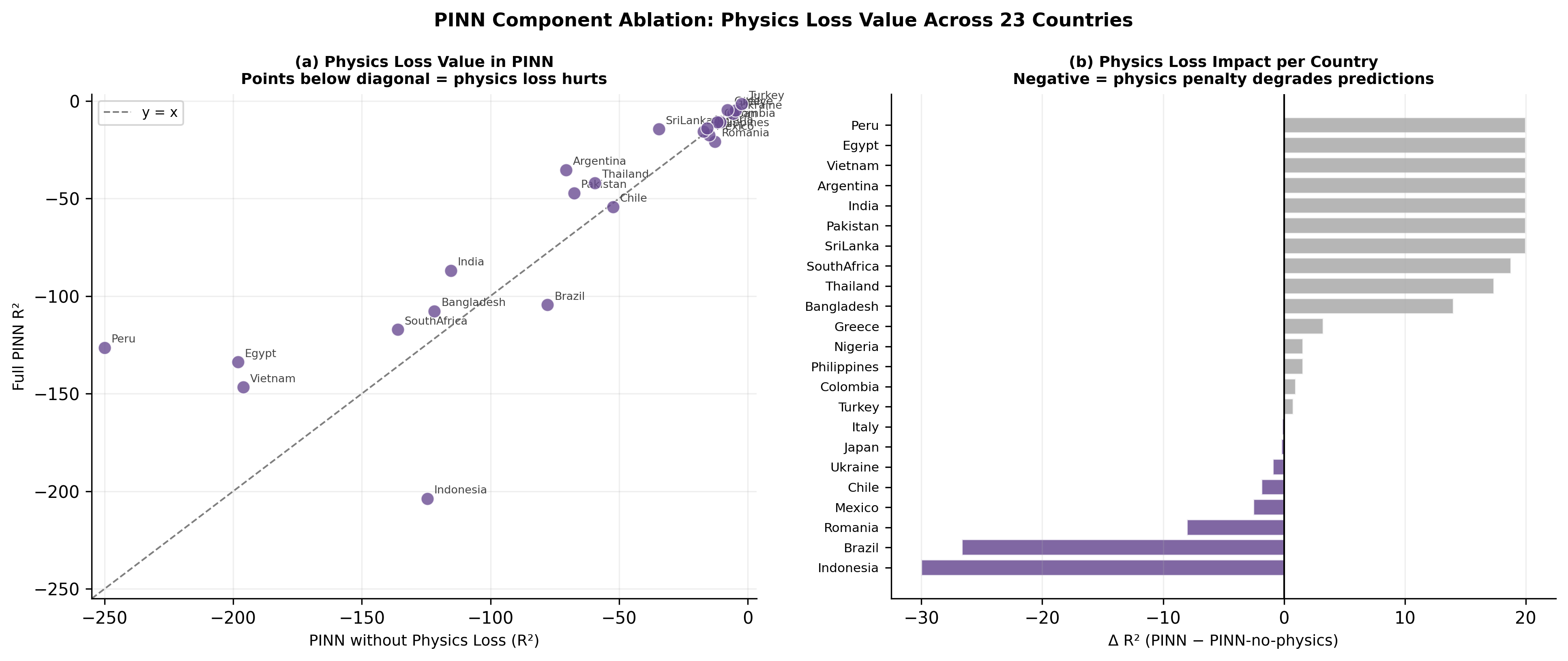}
    \caption{PINN ablation analysis. (a) Comparison of full PINN performance against a variant without the physics-informed loss term. Points below the diagonal indicate that removing the structural constraint improves performance. (b) Country-level change in test $R^2$ after removing the physics penalty. Negative values indicate degradation caused by the imposed heuristic prior.}
    \label{fig:pinn_ablation}
\end{figure*}

Figure~\ref{fig:pinn_ablation} shows that removing the structural penalty improves predictive performance in the majority of countries. In panel (a), most points lie below the diagonal, indicating that the full PINN performs worse than its unconstrained counterpart. Panel (b) quantifies the country-level impact, showing substantial degradation in several economies including Peru, Egypt, Vietnam, and Argentina.

This result provides direct evidence that the imposed heuristic structural priors are not merely ineffective, but actively harmful in many macroeconomic settings. Rather than regularizing learning toward meaningful dynamics, the physics penalty can introduce systematic bias when the assumed relationships do not align with observed economic behavior.

This strengthens the interpretation that prior misalignment, rather than model capacity alone, is a key explanation for the underperformance of structurally constrained approaches in this study.

\subsection{Failure Modes of Structural Priors}

Based on the empirical results, we identify four recurring failure modes that explain the behavior of models with structural priors:

\textbf{(1) Prior misalignment.} The imposed structural constraints do not accurately reflect the true macroeconomic relationships, leading to biased predictions.

\textbf{(2) Regime shifts.} Sudden changes in economic conditions violate assumptions of smooth dynamics, causing structured models to fail.

\textbf{(3) Structural breaks.} Discontinuities in the data, such as crises or policy shocks, are incompatible with the continuity assumptions embedded in the priors.

\textbf{(4) Optimization instability.} The interaction between data loss and prior loss introduces conflicting gradients, increasing sensitivity to initialization and training instability.

These failure modes provide a diagnostic framework for understanding when structural priors may harm rather than help model performance in non-physical domains.

\section{Conclusion}

This work presents a diagnostic study of scientific machine learning methods in macroeconomic forecasting, using a 23-country dataset as a stress-test environment. Across all models and countries, forecasting performance remains weak in absolute terms, with most $R^2$ values being negative. This highlights the inherent difficulty of modeling sparse, low-frequency, and non-stationary macroeconomic data.

Despite this, a consistent relative pattern emerges. Less-constrained models, particularly ARIMA and Neural ODE (NODE), outperform more-constrained approaches such as PINN and UDE across the majority of countries and temporal splits. The gap is especially pronounced in volatile and crisis-affected regimes. These findings suggest that the introduction of heuristic structural priors does not necessarily improve learning in practice, and can instead reduce both accuracy and stability when the assumed structure is not aligned with the data-generating process.

We interpret these results as a diagnostic rather than a general statement on the effectiveness of SciML. In domains where governing dynamics are uncertain, evolving, or partially observed, structural priors may act as misregularizers rather than useful inductive biases. This highlights the importance of empirically validating model assumptions and carefully selecting the level of imposed structure when applying SciML methods to real-world, non-physical systems.

\section{Acknowledgement}

The authors acknowledge the use of artificial intelligence (AI)-assisted tools for language refinement, grammar correction, and improving the readability of portions of this manuscript.

\bibliographystyle{plainnat}
\bibliography{references}

\appendix

\section{Appendix}
\paragraph{Extended Results}
The extended experimental results and supplementary visual analyses are presented on the following page.

\noindent\textbf{Note on country coverage.}
The primary analysis in the main paper is restricted to a standardized 23-country benchmark in order to ensure comparability across countries, indicators, temporal splits, and model evaluation procedures. During the broader experimental phase, we also evaluated additional countries, including Kenya and Poland. These results are reported in the appendix as exploratory supplementary analyses rather than as part of the core benchmark, because they were not used in the final standardized comparison presented in the main text. We retain them to demonstrate the broader scope of the experimental pipeline, while reserving the full expanded country set and additional visual analyses for a future journal-length version of this work.

\noindent\textbf{Data Sources.}
All country-level macroeconomic and development indicators used in this analysis were sourced from two publicly available databases.
Specifically:

\begin{itemize}
    \item World Bank World Development Indicators (WDI): \url{https://databank.worldbank.org/source/world-development-indicators}
    \item International Monetary Fund (IMF) Data: \url{https://data.imf.org/en}
\end{itemize}

\clearpage
\label{sec:appendix}

\begin{figure}[htbp]
    \centering
    \includegraphics[width=\linewidth]{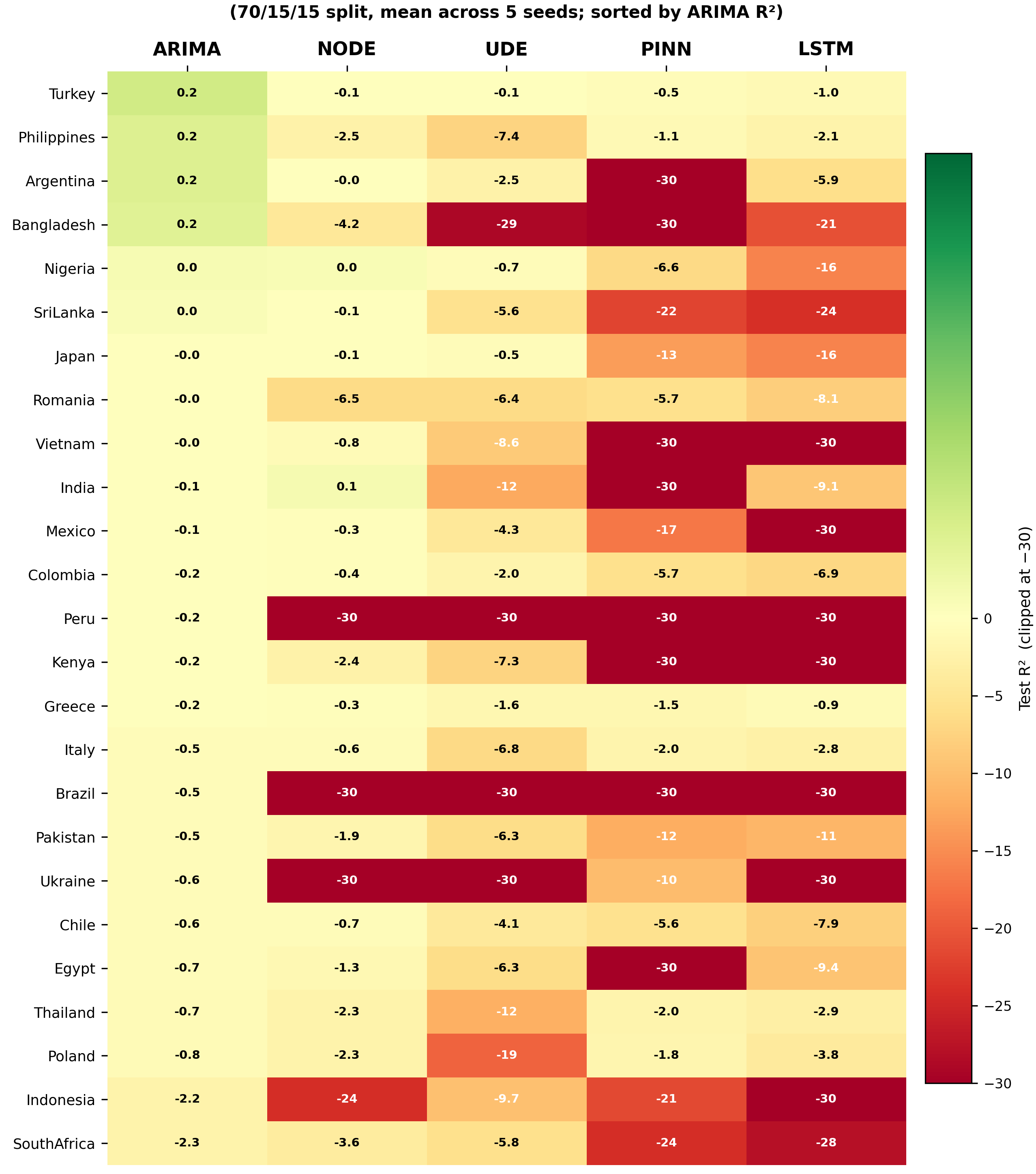}
    \caption{%
        \textbf{Full test R\textsuperscript{2} results across all 25 countries and five model families.}
        Results are shown for the 70/15/15 temporal split, averaged over 5 random seeds.
        Values are clipped at $-30$ for display; cells at $-30$ may represent substantially
        worse performance.
        Countries are sorted by ARIMA R\textsuperscript{2} in descending order.
        Green indicates R\textsuperscript{2} near zero or positive; red indicates severe
        underfitting relative to a mean baseline.
        ARIMA achieves the least-negative R\textsuperscript{2} in 24 of 25 countries.
        PINN exhibits extreme degradation in volatile economies such as Egypt, Bangladesh,
        and Kenya, while NODE is the second-best method in the majority of countries,
        achieving positive R\textsuperscript{2} in India and Nigeria.%
    }
    \label{fig:a1_r2_heatmap}
\end{figure}

\clearpage

\begin{figure}[htbp]
    \centering
    \includegraphics[width=0.9\linewidth]{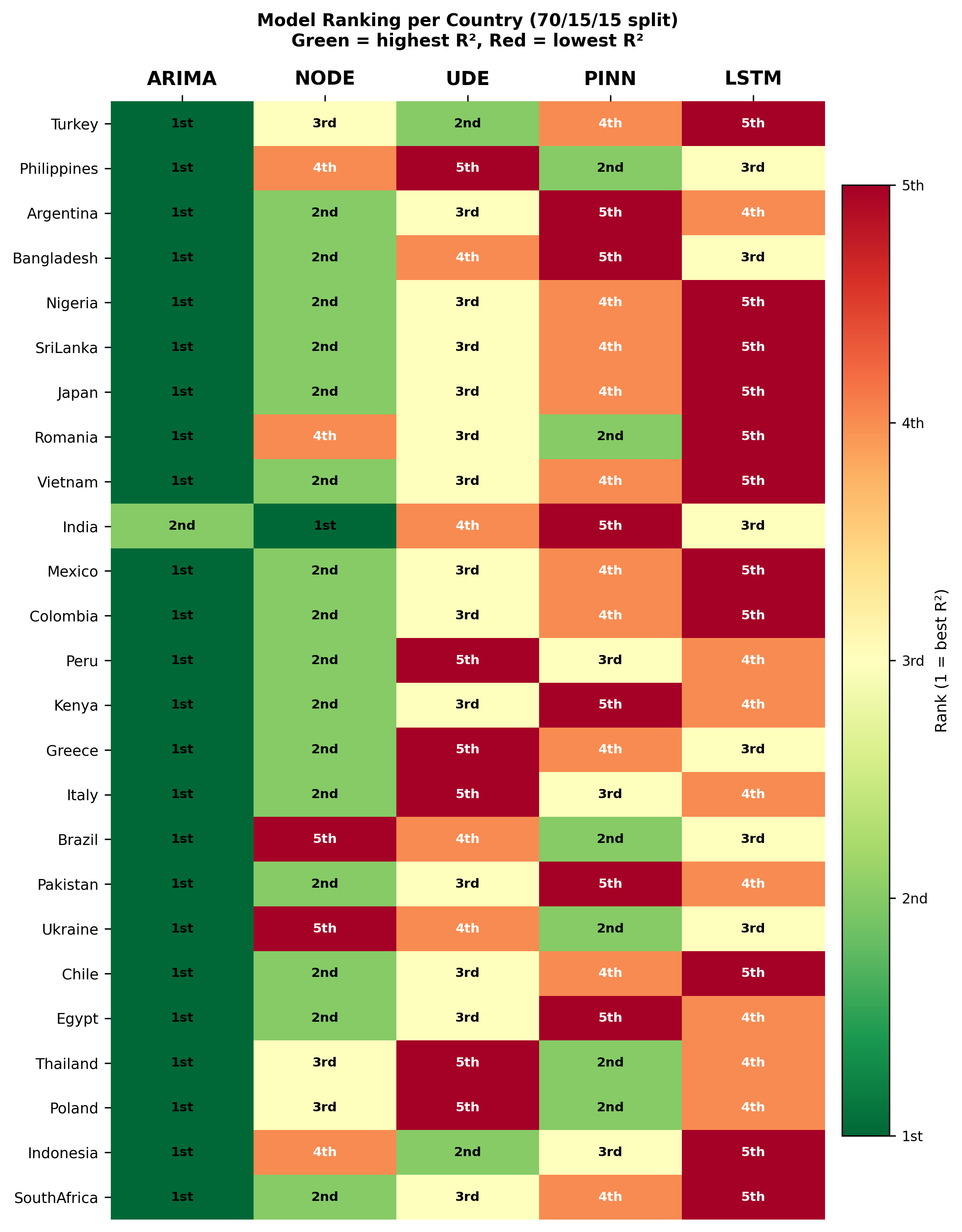}
    \caption{%
        \textbf{Per-country model ranking based on mean test R\textsuperscript{2}
        (70/15/15 split).}
        Rank~1 (green) indicates the highest R\textsuperscript{2}; Rank~5 (red) indicates
        the worst.
        The hierarchy is consistent across the 25 countries studied:
        ARIMA ranks first in 24 of 25 countries (NODE ranks first in India);
        NODE is ranked second in 16 countries;
        PINN and LSTM consistently occupy the bottom two positions, with LSTM
        ranking fifth in 15 countries and PINN ranking fourth or fifth in most countries.
        This near-total ordering across diverse economic regimes supports the
        interpretation that increased structural constraint systematically degrades
        forecasting performance in this domain.%
    }
    \label{fig:a2_rank_heatmap}
\end{figure}

\clearpage


\begin{figure}[htbp]
    \centering
    \includegraphics[width=\linewidth]{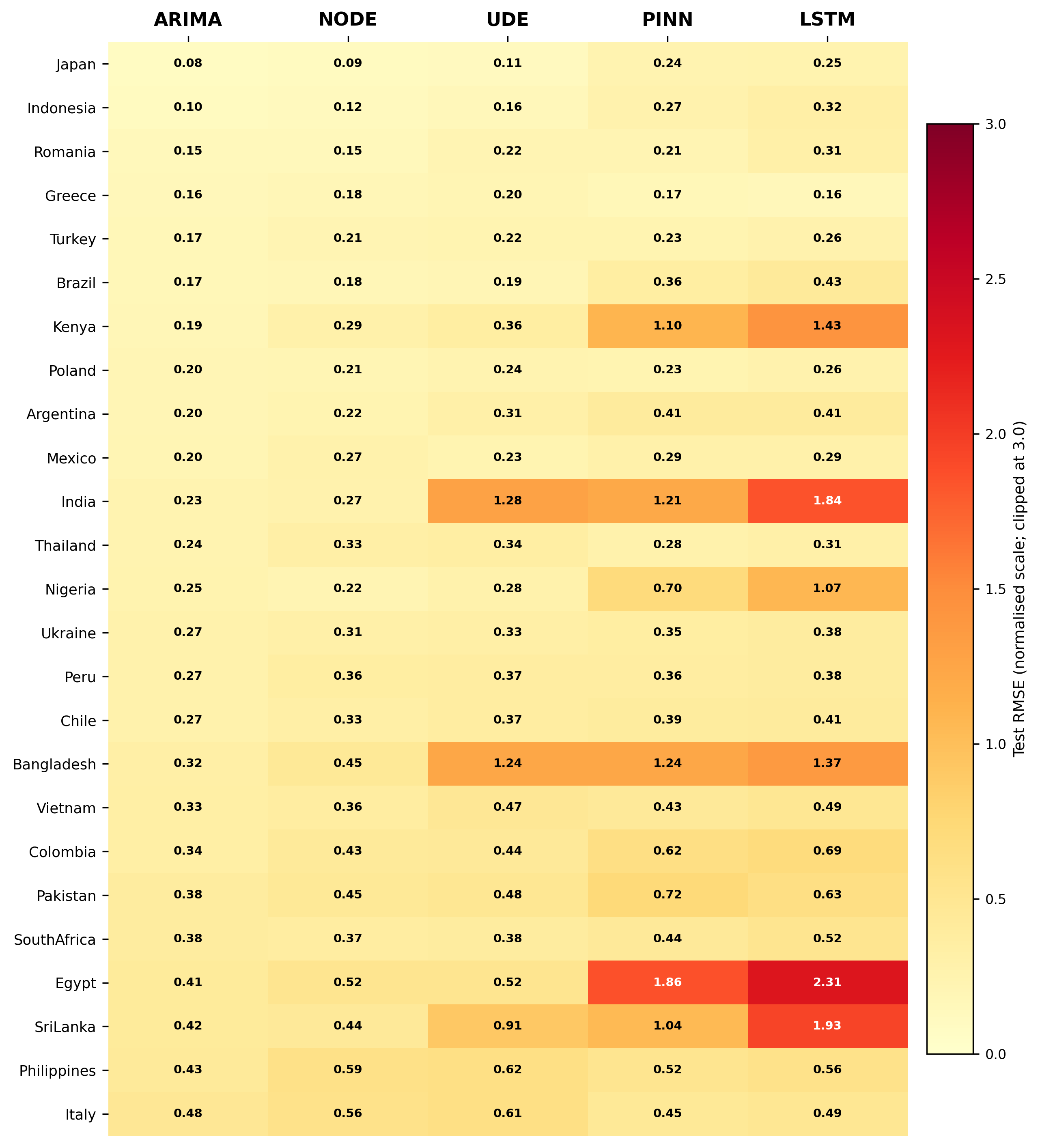}
    \caption{%
        \textbf{Full test RMSE results across all 25 countries and five model families.}
        Results are shown for the 70/15/15 temporal split, mean over 5 seeds, on the
        normalised feature scale.
        Values are clipped at $3.0$ for display; cells marked ``$>3$'' exceed this
        threshold.
        Countries are sorted by ARIMA RMSE in ascending order.
        Because R\textsuperscript{2} can become strongly negative when prediction
        variance is large relative to target variance—masking the absolute scale of
        errors—RMSE provides a complementary view of model quality.
        Under RMSE, ARIMA and NODE remain the consistently lowest-error methods, while
        LSTM and UDE show the highest absolute errors.
        PINN errors are moderate in absolute terms despite catastrophic
        R\textsuperscript{2}, reflecting the fact that PINN predictions track the range
        of the data but fail to capture its temporal structure.%
    }
    \label{fig:a8_rmse_heatmap}
\end{figure}

\clearpage

\begin{figure}[htbp]
    \centering
    \includegraphics[width=\linewidth]{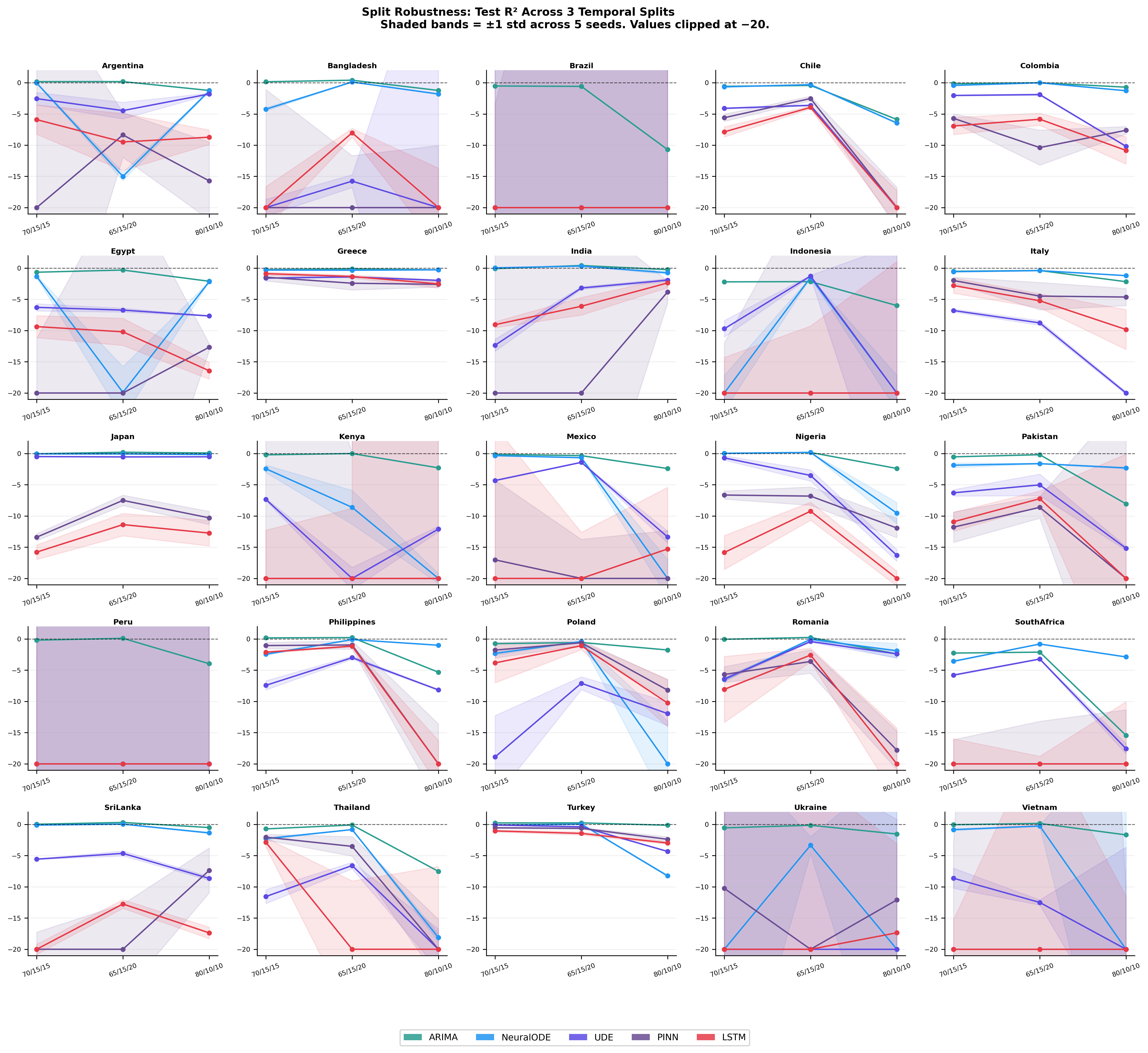}
    \caption{%
        \textbf{Split robustness: test R\textsuperscript{2} across three temporal split
        configurations for all 25 countries.}
        Split configurations are 70/15/15, 65/15/20, and 80/10/10.
        Shaded bands show $\pm 1$ standard deviation across 5 seeds; values are clipped
        at $-20$.
        The relative model hierarchy (ARIMA $\geq$ NODE $>$ UDE $>$ PINN $\geq$ LSTM)
        is preserved across splits in the large majority of countries, confirming that
        results are not an artefact of a particular training--test boundary.
        Countries such as India, Turkey, and Colombia exhibit stable ordering across
        splits, while Brazil, Peru, and Ukraine show high sensitivity consistent with
        structural breaks and regime shifts.%
    }
    \label{fig:a4_split_robustness}
\end{figure}

\begin{figure}[htbp]
    \centering
    \includegraphics[width=\linewidth]{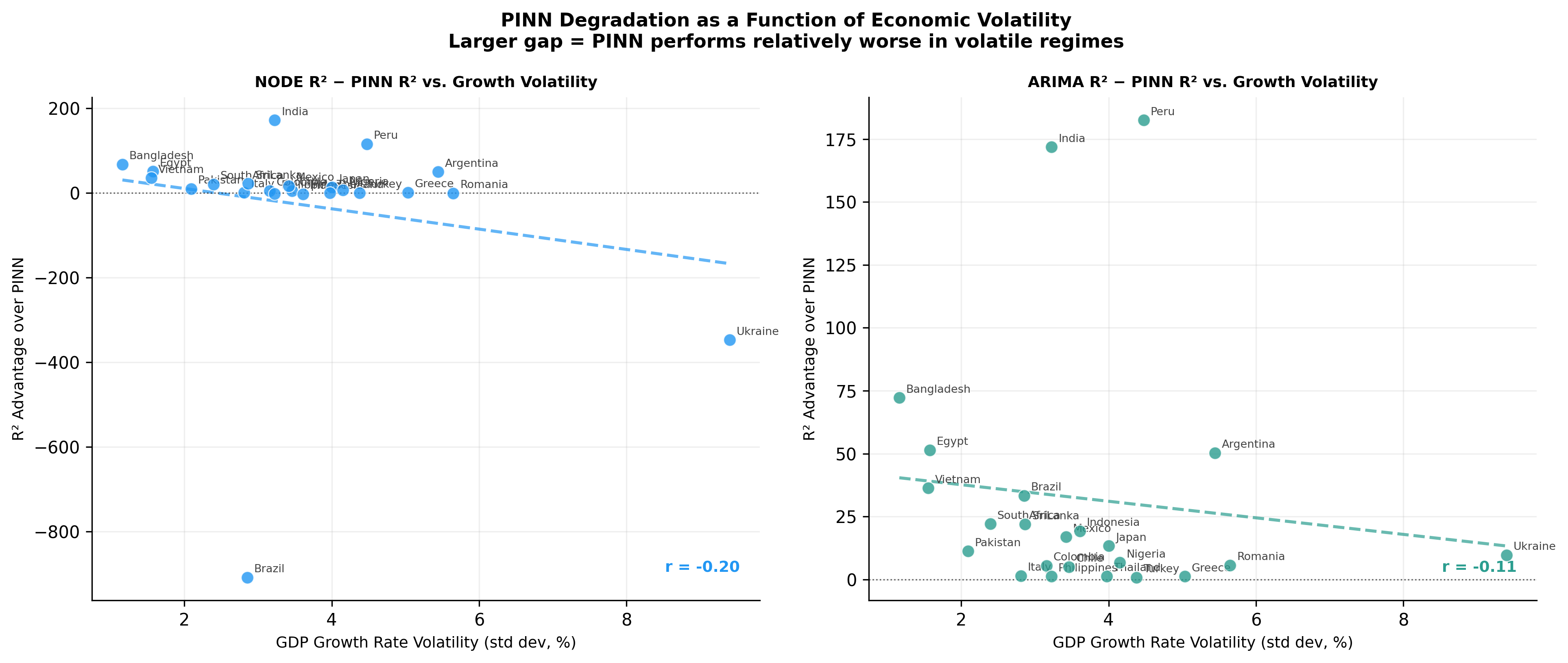}
    \caption{
    \textbf{PINN performance gap as a function of macroeconomic volatility.}
    Each point represents one country. The $y$-axis reports the test $R^2$ advantage of NODE (left) and ARIMA (right) over PINN, so positive values indicate that the reference model outperforms PINN. The relationship between GDP growth volatility and the PINN performance gap is weak in both panels ($r=-0.20$ for NODE--PINN and $r=-0.11$ for ARIMA--PINN), suggesting that volatility alone does not fully explain PINN degradation. Instead, the large positive gaps in several countries indicate that PINN underperformance is likely driven by a combination of prior misalignment, structural breaks, and optimization instability rather than by growth volatility alone.
}
    \label{fig:a5_pinn_volatility}
\end{figure}

\begin{figure}[htbp]
    \centering
    \includegraphics[width=\linewidth]{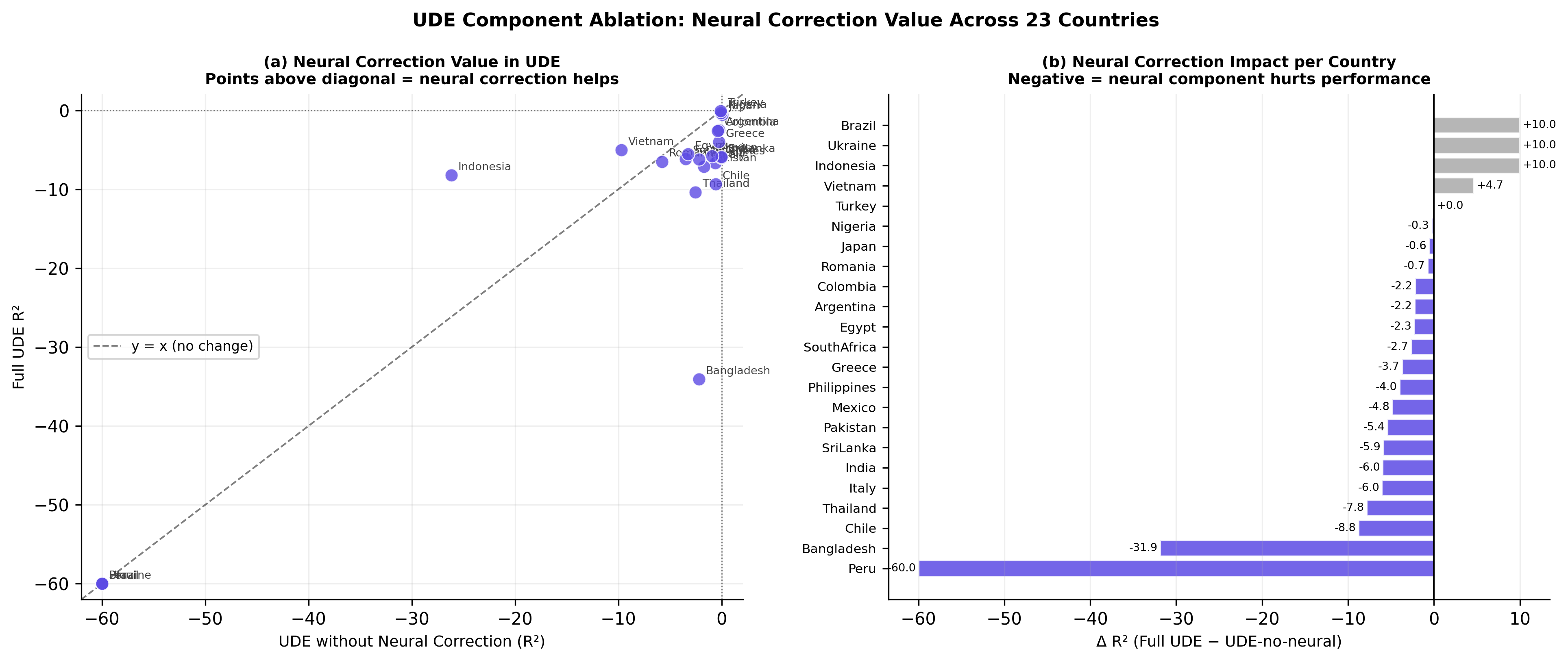}
    \caption{%
        \textbf{UDE component ablation: neural correction value across 23 countries.}
        \emph{(a)}~Scatter of full UDE R\textsuperscript{2} vs.\ UDE without the learned
        neural correction term. Points below the diagonal indicate that removing the
        neural correction \emph{improves} performance.
        \emph{(b)}~Per-country $\Delta$R\textsuperscript{2}~$=$~(Full UDE)~$-$~(UDE-no-neural).
        Negative values (purple) confirm the neural correction is detrimental in 18 of
        23 countries.
        This suggests the learned correction does not meaningfully reduce the residual
        error left by the structural ODE terms; instead it introduces additional fitting
        noise amplified by the already misspecified physics prior.
        Brazil, Ukraine, and Indonesia exhibit the largest negative impact.%
    }
    \label{fig:a6_ude_ablation}
\end{figure}


\begin{figure}[htbp]
    \centering
    \includegraphics[width=\linewidth]{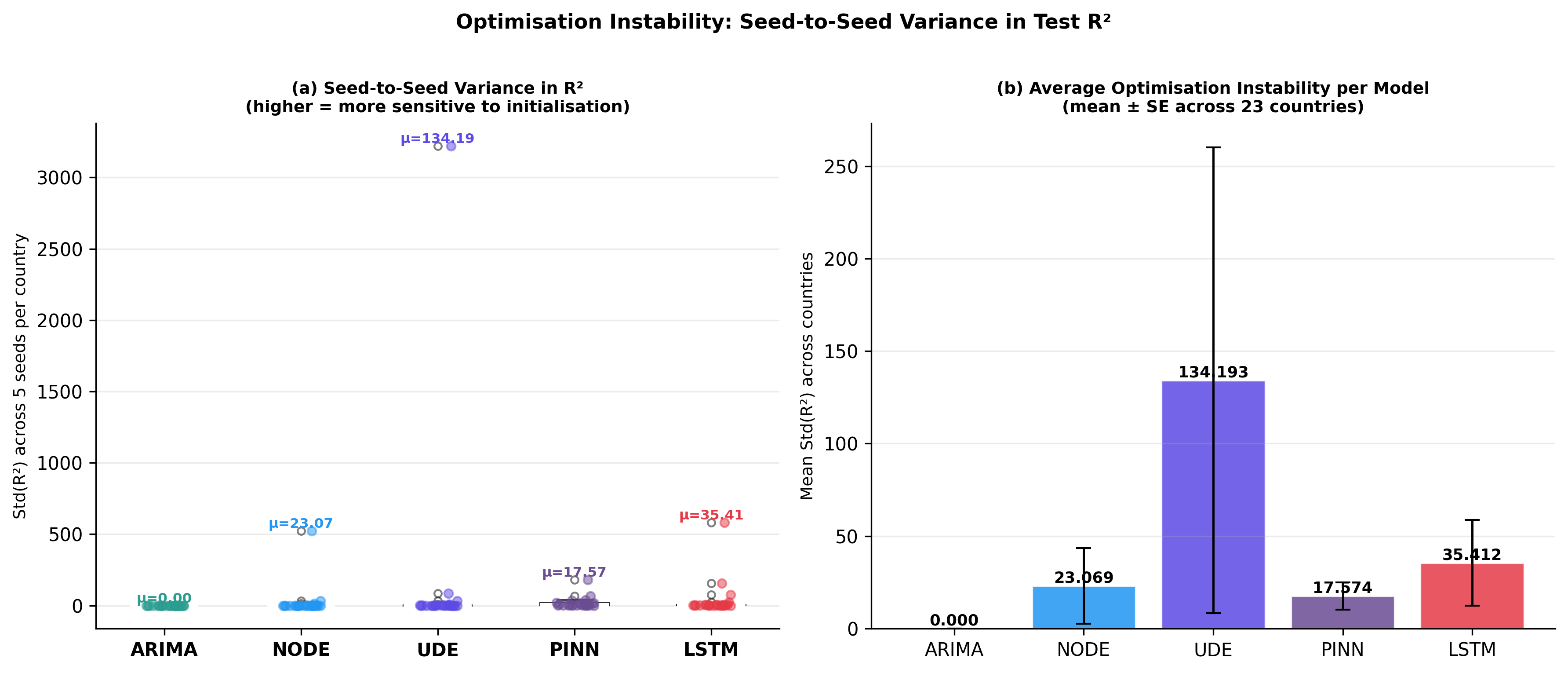}
    \caption{
    \textbf{Optimisation instability across random seeds (70/15/15 split, 23 countries).}
    (a) Distribution of per-country test $R^2$ variability across five random seeds for each stochastic model, measuring sensitivity to initialization.
    (b) Mean optimisation instability across all 23 countries, with error bars indicating uncertainty across countries.
    ARIMA is deterministic and therefore exhibits zero variability.
    UDE shows the highest instability ($\mu = 134.2$), driven by catastrophic divergence in countries such as Peru, Brazil, and Ukraine.
    NODE is substantially more stable ($\mu = 23.1$), while LSTM shows intermediate instability ($\mu = 35.4$) and PINN remains comparatively lower ($\mu = 17.6$).
    The pronounced instability of UDE suggests that hybrid structural priors can introduce conflicting optimisation signals, making training substantially less stable.
    }
    \label{fig:a3_instability}
\end{figure}

\end{document}